\documentclass[10pt, conference, compsocconf]{IEEEtran}
\ifCLASSINFOpdf
\else
\fi

\usepackage{graphicx}
\begin{document}
%
\title{D-OccNet: Detailed 3D Reconstruction Using Cross-Domain Learning}




%
\author{\IEEEauthorblockN{Minhaj Uddin Ansari\IEEEauthorrefmark{1},
Talha Bilal\IEEEauthorrefmark{2},
Naeem Akhter\IEEEauthorrefmark{3}}
\IEEEauthorblockA{\IEEEauthorrefmark{1}School of Computing\\
Queen's University,
Kingston, Ontario K7M2W8\\ Email: 20mua@queensu.ca}
\IEEEauthorblockA{\IEEEauthorrefmark{2}Escuela Politécnica Superior (EPS)\\
Universidad Autónoma de Madrid,
Madrid, Spain 28049\\
Email: talha.bilal@estudiante.uam.es}
\IEEEauthorblockA{\IEEEauthorrefmark{3}Department of Computer and Information Sciences\\
Pakistan Institute of Engineering and Applied Sciences,
Islamabad, Pakistan 45650\\
Email: naeemakhter@pieas.edu.pk}}


\maketitle

\begin{abstract}
Deep learning based 3D reconstruction of single view 2D image is becoming increasingly popular due to their wide range of real-world applications, but this task is inherently challenging because of the partial observability of an object from a single perspective. Recently, state of the art probability-based Occupancy Networks reconstructed 3D surfaces from three different types of input domains: single view 2D image, point cloud and voxel. In this study, we extend the work on Occupancy Networks by exploiting cross-domain learning of image and point cloud domains. Specifically, we first convert the single view 2D image into a simpler point cloud representation, and then reconstruct a 3D surface from it. Our network, the Double Occupancy Network (D-OccNet) outperforms Occupancy Networks in terms of visual quality and details captured in the 3D reconstruction. 

\end{abstract}


%
\IEEEpeerreviewmaketitle

\section{Introduction}
3D image reconstruction from single view 2D images is a fundamental challenge in computer vision. In recent days, demand for 2D to 3D reconstruction has increased due to its wide range of applications in autonomous vehicles, virtual reality, medicine, augmented reality and robotics. Although this task is the basic functionality of the human brain, it is a very complex problem for a computer. Fortunately, with the availability of large open source 3D datasets like ShapeNet [16], deep learning methods have gained rapid advancement in this area. Deep learning algorithms are now capable of developing 3D images in a variety of domains due to supervised learning on 3D datasets. Presently, deep learning algorithms for reconstructing 3D images are divided into three major categories: voxel, point cloud and mesh  based on output domain representation of the reconstructed image. We will briefly go through the research contributions in all three groups. 

Voxels are small cubic units depicting pixels in 3D domain, where each individual pixel has a length, width, and height. In recent years, there has been a lot of work done in voxel domain due to its simplicity. However, since each voxel must occupy volume, 3D images in the voxel domain take up a lot of memory and have coarse boundaries and surfaces, inevitably resulting in lower resolutions. C.B Choy et al [1] introduces 3D Reconstruction Neural Network (3D-R2N2) which is an extension of the standard Long Short Term Memory (LSTM) network. 3D-R2N2 unifies single-view and multi-view 3D reconstruction in a single framework and takes the advantage of the power of the LSTM to retain previous observations and incrementally refine the output reconstruction as more observations become available. H. Xie et al [17] uses context aware fusion module in its encoder-decoder structure for high quality reconstruction in voxel domain. X. Yan et al [18] also uses encoder-decoder architecture and introduces a 2D silhouette loss function based on perspective transformations. J. Wu [15] leverages the power of 3D convolutional networks and generative adversarial nets to accomplish this task of 3D reconstruction in an adversarial fashion. MarrNet [14] is an end to end trainable model consisting of a 2.5D sketch estimator, a 3D shape estimator and a reprojection consistency function to enforce alignment between estimated 3D shape and inferred 2.5D sketches. 

\begin{figure}[h!]
\includegraphics[width=0.95\columnwidth]{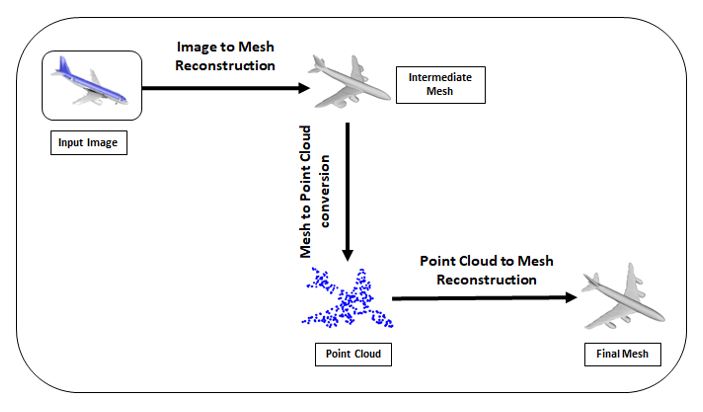}
\centering
\caption{ Overview of the D-OccNet }
\end{figure}

Point clouds are points distributed throughout a bounded 3D space. Since the points are distributed with a high degree of freedom, and there aren't any connections between them to ensure they are in the proper position relative to one another, it is difficult to control the 3D images in the point cloud domain generated by deep learning algorithms. H. Fan et al [2] introduces a novel conditional shape sampler, capable of predicting multiple plausible 3D point clouds from an input image. P. Mandikal et al [10] proposes an latent embedding matching approach for 3D reconstruction. They first train a 3D point cloud auto-encoder and then learn a mapping from the 2D image to the corresponding learnt embedding. They introduced a probabilistic training scheme in the image encoder to obtain multiple plausible 3D mappings from a single input image. C. Lin et al [8] uses 2D convolutional operation based structure generator to predict 3D structure and introduces a pseudo-rendering pipeline to serve as a differentiable approximation of true rendering. They also use the pseudo-rendered depth images for 2D projection optimization for learning dense 3D shapes.  Zhao et al [19] introduced Variational Auto Encoder (VAE) based 3D Point Capsule Network architecture to process sparse 3D point cloud.

A mesh a set of vertices and edges connected to form surfaces. Meshes are the best choice for 3D images because they aren't as memory intensive as voxel images, they have smoother surfaces compared to voxel images, and the vertices are connected by edges to one another, so the structure and texture rendering is more stable compared to point cloud images. C. Kong et al [6] uses dictionary of dense CAD models and propose a two-step strategy. First it employs orthogonal matching pursuit to rapidly choose the closest single CAD model in the dictionary to the projected image. Second, it employs a novel graph embedding based on local dense correspondence to allow for sparse linear combinations of CAD models. T. Groueix et al [3] introduces a novel 3D surface generation network AtlasNet, which is composed of a union of learnable parametrizations. These learnable parametrizations transform a set of 2D squares to the surface. A recent trend is to use deformation network for 3D reconstruction. D. Jack et al [5] introduces a network that learns to deform points sampled from a high-quality mesh and produces meshes with fine-grained geometry. Pixel2Mesh [13] uses graph convolutions to deform an initial ellipsoid according to features extracted from the images. This technique is really limited to reconstruct a few categories because it highly depends upon shape of initial 3D shape (ellipsoid). 

Many techniques in Computer Vision focus on 3D reconstruction from more than one perspective (multiple 2D views) of an object, but in many cases we are only available with one perspective (single 2D view) of the object. 3D reconstruction using single view is more difficult task because you are only provided with partial observation of object which can lead to infinite representations. However it is also the most natural considering humans can easily visualize the 3D representation of an object by looking at it from a single perspective. In this paper we focus on 3D reconstruction from only a single view of the object.

Our Double Occupancy Network (D-OccNet) is inspired by the state of the art Occupancy Networks (OccNet) [11] that learns boundary of object in 3D space using occupancy values. The difference between OccNet and D-OccNet is that OccNet only uses a single domain for image to mesh conversion, whereas D-OccNet is based on cross domain model learning. In D-OccNet, the input 2D single view image is first transformed into a point cloud and then post-processor transforms the point cloud into the mesh domain. This two-step transformation helps better reconstruct an image in the 3D domain. While quantitative results of D-OccNet are comparable to OccNet, it outperforms OccNet in qualitative comparison.

The rest of this paper is organized as follows: Section 2 describes the working of Occupancy Network; Section 3 explains the architecture of learning based D-OccNet model; Section 4 discusses the training setup and experimental results; and, finally, Section 5 concludes this work.

\section{Occupancy Networks}

The Occupancy Network (OccNet) is an encoder decoder convolution neural network architecture. The encoder encodes the input into a vector and the decoder generates a 3D mesh from that vector. Occupancy Networks can convert 2D single view images, point clouds and voxels to probability occupancy maps and these maps are used to extract corresponding 3D meshes. For different types of inputs, different types encoders are used in the Occupancy Network; however, in all cases, the decoder remains the same.

\subsection{Encoder}
As discussed in the previous section, an occupancy network can encode either an image, or a point cloud or a voxel. A 2D classification networks encoder, such as ResNet18 [4] can be used if the input is a single view 2D image. A PointNet [12] encoder is used if the input is a point cloud and a 3D convolution encoder network is used if the input is a voxel. When a 2D single view image is passed to ResNet18; a point cloud is passed to a PointNet; or a voxel is passed to a 3D convolution encoder network, the encoder outputs a vector. The dimensions of the vector can vary, but the decoder can be adjusted to learn the features encoded in the vector regardless of its dimensionality. 

\subsection{Decoder}
The decoder uses CBN (Conditional Batch Normalization), fully connected layers and Resnet blocks to generate an occupancy map from the input encoded vector. Nx3 3D-coordinate points are passed into a fully connected layer to generate a feature vector of Nx256 dimensions where each point is associated with a 256-dimensional vector. This feature vector is then passed into a Resnet block. The Resnet block starts with applying CBN layer, followed by ReLU activation and then a fully connected layer. The output is sent into a second CBN layer, followed by ReLU activation and a second fully connected layer. The output is then added to the Resnet input. This process repeats five times. After five iterations, the output is again passed through a CBN and ReLU layer, and finally a fully connected layer reduces the output to a Nx1 1-dimensional vector. This vector is passed through a sigmoid function to produce the occupancy probabilities corresponding to each point in the input Nx3 vector. The CBN layer receives the encoded vector c from the encoder and passes it through two fully connected layers $\beta$(c) and $\gamma$(c). The input feature vector is normalized using first and second order moments, multiplied with the output of fully connected layer $\gamma$(c) and then added to the output of fully connected layer $\beta$(c). In order to calculate occupancy value, cross entropy loss is used which is calculated on each minibatch. 

\[ f_{out}=\gamma(c){(f}_{in}-\mu)/\sqrt{\sigma^2+\epsilon}+\beta(c) \]

\subsection{Mesh Extraction}
The decoder outputs the occupancy values associated with each point in the Nx3 3D-coordinate points matrix initially sent as input. Points that have high occupancy value, or high probability, are considered lying inside the object and points that have lower occupancy values are considered lying outside the object. Multiresolution Iso-Surface Extraction (MISE) [7][9] is used to develop cube around these points. The cubes can then be broken down to the desired resolution. After the desired resolution is obtained, the marching cube algorithm [20] is used to convert the voxel to a mesh. 

\section{Method}

We found that 2D image to 3D mesh conversion by the OccNets resulted in meshes that weren’t as perfect compared to their corresponding ground truths. This is because the Occupancy Network encoder CNN could only generate latent vectors containing information of the 2D image fed to it. There wasn't any 3D information embedded into the latent vector. We hypothesized that if we used a point cloud representation of the input image, thereby integrating some 3D information, and fed it into the Occupancy Network encoder, the latent vector produced by the encoder would contain some 3D information of the input, resulting in better quality 3D meshes. Using this hypothesis, we introduced Double Occupancy Network (D-OccNet) as an extension of OccNet. The "D" in D-OccNet refers to the architectural design of this network. It contains two OccNets connected together by a small processing unit in between the output of the first OccNet and the input of the second OccNet. The main job of this small processing unit is to convert the mesh output of the first OccNet into a suitable point cloud input for the second OccNet. We will discuss the architecture in detail in the upcoming section. 

\begin{figure}[h!]
\includegraphics[width=0.95\columnwidth]{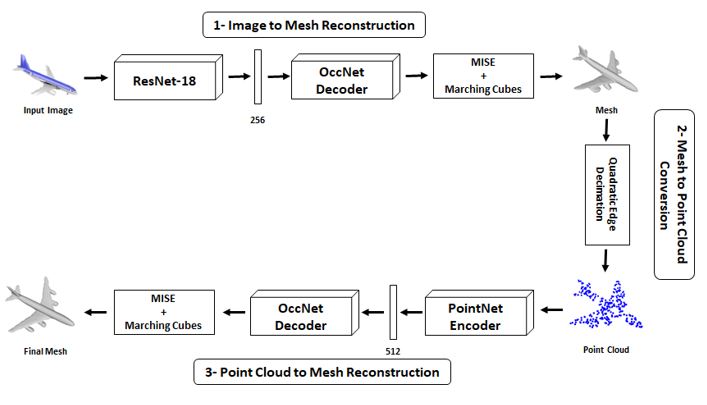}
\centering
\caption{ Internal architecture of D-OccNet }
\end{figure}

D-OccNet contains three steps in the following sequence (1) image to mesh reconstruction, (2) mesh to point cloud conversion and (3) point cloud to mesh reconstruction.

\subsection{Image to Mesh Reconstruction}
A single view 2D image is fed into a pre-trained Resnet18 model. Resnet18 encodes the image into a latent vector of size 256x1. Resnet18 can be replaced with any 2D CNN network to generate a latent vector. This latent vector is sent to the OccNet’s decoder to produce a 3D mesh.

\subsection{Mesh to Point Cloud Conversion}
The 3D mesh generated in the first step contains approximately 8000-12000 vertices connected by edges. This is as far as the original OccNet goes. The PointNet encoder has been trained on 300 points point cloud as input. To make the input suitable for PointNet, we use MeshLab, an open source 3D mesh processing software to reduce the number of points in the 3D mesh to 300. We save the 300 points of the generated mesh as a point cloud, thereby removing the edge information.

\subsection{Point Cloud to Mesh Reconstruction}
The point cloud from the previous step is fed into the PointNet encoder. PointNet generates a latent vector of size 512x1, which is sent into the OccNet's decoder. The decoder generates a 3D mesh from the latent vector. This 3D mesh is the final output of our D-OccNet. 

\section{Training and Results}

\begin{figure}[h!]
\includegraphics[width=0.93\columnwidth]{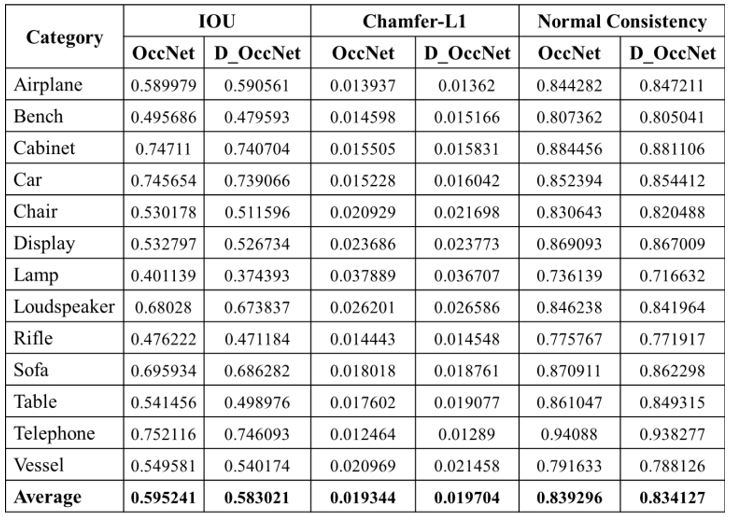}
\centering
\caption{ Comparison of IoU, Chamfer and Normal Consistency between OccNet and D-OccNet }
\end{figure}

In our implementation of D-OccNet, the training of two OccNets are involved. We train the first Occnet to generate meshes from single view 2D images. The first OccNet contains a ResNet18 encoder and OccNet’s decoder. We then train our second OccNet to generate meshes from point cloud.The second OccNet contains a PointNet encoder and the OccNet’s decoder.

We use the same dataset as as used by the authors in [1]. This dataset contains 13 object categories of 3D CAD models taken from the ShapeNet dataset. ShapeNet is a large annotated dataset that contains 12,000 CAD models belonging to 270 different object categories. It is one of the most widely used dataset for training supervised deep learning models. 

We augment the input data by scaling it by a random factor between 0.75 and 1 and then taking a random crop. Train/Test split ratio is set the same as in [2]. Batch size is set to 64 and the Adam optimizer is set with a learning rate of 1e-4, $\beta$1 = 0.9 and $\beta$2 = 0.999. Validation accuracy is set as the stopping criteria for training. The code has been written in Python2. For the training, we use NVIDIA GTX 1070 GPU in a 64 GB RAM Core i7 computer running on Ubuntu OS.

We compare the output meshes from D-OccNet with the original OccNet. While surface generation methods struggle to capture shape details such as table surface, airplane wings and car wheels, our method preserves these details that reside in the source mesh. We evaluate D-OccNet’s output both quantitatively and qualitatively and draw a comparison with the original OccNet. For Quantitative comparison we use IOU, Chamfer-L1 and Normal Consistency metrics. IoU is the intersection over union metric which computes the overlapping of two meshes, in our case the ground truth and predicted mesh, with the union. If the overlapping is same, it will give a score closer to 1, meaning the final output is closer to the ground truth. As evident by the Fig. 4, our D-OccNet produces meshes having IOU comparable to the original OccNet. The Chamfer loss compares the distance between points in the output and predicted meshes. The smaller the distance, the lesser the dissimilarity is between the meshes.  Again, our meshes outperform OccNet’s meshes when Chamfer L1 losses are compared. Finally, Normal Consistency metric shows that the normal of our mesh are closer to the normal of the ground truth mesh.

Quantitatively D-OccNet is comparable to OccNet but our meshes have visual superiority. Figure 5 shows a visual comparison between the outputs of both networks. In case of the airplane mesh, our D-OccNet reconstructed the front wheel of the plane, which wasn’t reconstructed by OccNet. The engine fans are also much smoother and sharper. The table mesh has sharper edges and finer corners, and the car mesh has sharper passenger seats and includes rearview mirrors. 

The main reason for similar quantitative result was due to choice of metrics. There is no available metric that can differentiate the little details between two meshes. For example, if OccNet generates a 3D mesh of a car that has no rearview mirrors but is of the right volume,  and if D-OccNet produces the same car mesh with rearview mirrors, but the car is slightly larger in volume than the ground truth, the IoU metric values obtained by comparing the ground truth of both networks will be similar because the volume overlapping with the ground truth is almost identical in both cases. In other words, the IoU of OccNets mesh output missing the rearview mirror is same is the IoU of D-OccNets mesh output having the rearview mirror but the body being larger in size. Although this might be critical in medical applications where small changes of volume can hold great significance, other applications such as recreating an environment based on the images captured by a self driving car camera or creating a hologram of a product from an online shopping store might not need this level of precision. Instead, the greater focus in these type of applications may be to correctly reproduce the image visually. A novel metric is needed to capture this difference.

OccNet’s performance was not good in reconstructing 3D objects from single view 2D images because it did not get any 3D information about the input. We exploited this draw back by first generating 3D information about the image in the form of point clouds and then using this 3D information for reconstruction purposes which significantly improved our output quality.

\begin{figure}[h!]
\includegraphics[width=0.95\columnwidth]{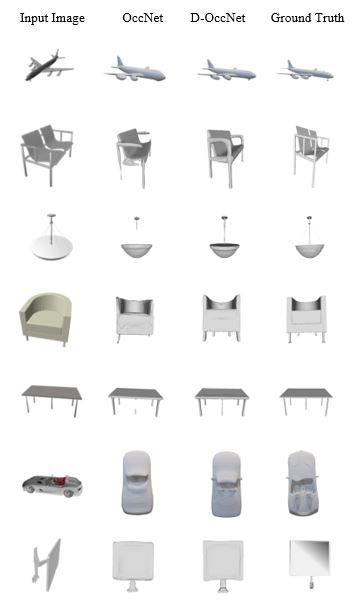}
\centering
\caption{ Qualitative comparison between OccNet and D-OccNet  }
\end{figure}

\section{Conclusion}
In this work, we designed a method for a detailed and precise 3D reconstruction from a single 2D perspective of an object. Our three-step network converted a 2D image into a point cloud, and then created a 3D object using the point cloud as input. We showed that D-OccNet could generate better 3D objects, in terms of visual detail compared to the current state of the art method. 

\section*{Acknowledgment}

This research primarily took place in Pakistan Institute of Engineering and Applied Sciences, Pakistan. Minhaj Uddin Ansari and Talha Bilal were both undergraduate students at that time. Minhaj is now pursuing masters in Queen’s University, Canada. Talha Bilal is now pursuing masters under the Erasmus program. We (Minhaj and Talha) are grateful to Pakistan Institute of Engineering and Applied Sciences for providing us the resources, and for the continuous support and motivation given to us by our supervisor Dr. Naeem Akhter.



%
\nocite{*}
\bibliographystyle{IEEEannot}
\bibliography{annot}

\end{document}